\documentclass[letterpaper, 10 pt, conference]{ieeeconf}  % Comment this line out if you need a4paper

\IEEEoverridecommandlockouts                              % This command is only needed if 
\usepackage{amsmath,amssymb}
\usepackage{amsfonts}
\usepackage{soul}
\usepackage{bm} % \bm{} command for bold math arguments
\usepackage[cal=cm,bb=ams,frak=mma,scr=euler]{mathalfa}
\usepackage{graphicx}
\usepackage[table,usenames,dvipsnames]{xcolor} % has to come before tikz
\usepackage{pgfplots}
\usepackage{tikz,tikzscale}
\usetikzlibrary{shapes,shapes.geometric,arrows,patterns,automata,positioning,calc}
\usepackage{subfig}
\usepackage{multirow}
\usepackage{mdframed}
\usepackage{cite}
\usepackage{algorithm,algpseudocode}
\usepackage{array}
\usepackage{hyperref}

\usepackage{amsmath}
\DeclareMathOperator*{\argmax}{arg\,max}

\pgfplotsset{compat=1.16}

\usepackage{latexsym}
\usepackage{mathtools}
\usepackage{relsize}
\usepackage{xcolor}
\usepackage[dvipsnames]{xcolor}
\usepackage{multirow}
\usepackage{graphics} % for pdf, bitmapped graphics files
\usepackage{epsfig} % for postscript graphics files
\usepackage{mathptmx} % assumes new font selection scheme installed
\usepackage{times} % assumes new font selection scheme installed
\usepackage{amsmath} % assumes amsmath package installed
\usepackage{amssymb}  % assumes amsmath package installed
\newtheorem{Rem}{Remark}
\newtheorem{lemma}{Lemma}

\title{\LARGE \bf
Communication-free Cohesive 
Flexible-Object 
Transport using \\
Decentralized Robot Networks
}

\author{Yoshua Gombo, Anuj Tiwari and Santosh Devasia% <-this % stops a space
\thanks{Yoshua Gombo, Anuj Tiwari and Santosh Devasia are with the Department of Mechanical Engineering,
        University of Washington, 3900 E Stevens Way NE, Seattle, WA 98195.
        {\tt\small ygombo@uw.edu}}%
}

\begin{document}

\maketitle

\thispagestyle{empty}
\pagestyle{empty}

%%%%%%%%%%%%%%%%%%%%%%%%%%%%%%%%%%%%%%%%%%%%%%%%%%%%%%%%%%%%%%%%%%%%%%%%%%%%%%%%
\begin{abstract}
% Current {\color{green} 
Decentralized network theories focus on achieving consensus and in speeding up the rate of convergence to consensus. However, network cohesion (i.e., maintaining consensus) during transitions between consensus values  is also important when  transporting flexible structures. Deviations in the robot positions  due to loss of cohesion when moving flexible structures from one position to another, such as uncured-composite aircraft wings, can cause large deformations, which in turn, can result in potential damage. The major contribution of this work is to develop a decentralized approach to transport flexible objects in a cohesive manner 
using local force measurements, without the need for additional communication between the robots. Additionally, stability conditions are developed for discrete-time implementation of the  proposed cohesive-transition approach, and   experimental results are presented, which show that the proposed cohesive 
transportation approach can reduce the relative deformations by $85\%$  when compared to the case without it.
\end{abstract}
%
% Index terms (keywords) options: Agent-based systems, control of networks, cooperative control, decentralized control, flexible structures, transportation networks
%
\section{Introduction}
The main goal is to transport flexible objects cohesively (i.e., all robots move in a similar manner) using decentralized robot networks. 
Network control theories can be used to  rapidly transition from one equilibrium (where all the robots in the network are in consensus) to another, i.e., a new consensus value, which also can be applied for transporting an object using robot networks. However, current network theories focus on the speed of convergence to a new consensus value~\cite{Nesterov_83,VanScoy_Lynch_18}, and they do not aim to ensure that the robot responses remain cohesive during transition.
For a transport task, the lack of cohesion during transition can lead to large deformation and cause damage to the object being transported. 
While centralized communication can yield to low-deformation transport~\cite{KarHan1996, Khoshnevis1998}, there is a growing interest in decentralized transport using a robot network that only uses local sensing due to robustness to one or more robot failures~\cite{Petitti2016, Tsiamis2015}, adaptability to varying number of robots~\cite{WangSchwager2016, jackson_2020}, and versatility to transport different objects \cite{Petitti2016, chen_2015}, without the need for centralized control and communication~\cite{Montemayor2005}.
The main contributions of this paper are to propose a cohesive transport approach for flexible structures with decentralized robot networks, and to establish stability conditions for a discrete-time 
implementation using local force sensing.
\vspace{0.1in}
Observation in nature indicates that ants seem to use local force measurement in their movement to transport foods, rather than communicating explicitly, e.g.,\cite{Gelblum2015}.
% {\color{blue}It is observed that ants sense a small vibrations and deformations on the food in order to move in consensus \cite{McCreery2014}}.
Similarly, in a transport task the elasticity of the object can be used to transmit information (i.e., forces and positions) among neighbors in network, instead of communicating with each other.
For example, as shown in \cite{BaiWen2010, loianno_2018, WangSchwager2016, Thapa2019}
measurements of the local force  exerted between the flexible object and robot can be used to infer the local deformation and accomplish the transport task in a decentralized manner. Alternatively, changes in the desired shape of the object can also be measured to develop a decentralized feedback control for object transport, e.g., \cite{MoraKnepperSiegwartRus2015,flixeder_2016, rossi2020}.
While such methods can be used to achieve transport of flexible objects from one position to another, say within some specified settling time, there is no direct control over the resulting deformations on the object. 
In the presence of only a  few leaders (who have access to desired transport  trajectory), there can be substantial 
deviation in the robot positions away from the leaders resulting in distortion and potential damage. For example when transporting uncured-composite aircraft wings, large deformations  can lead to structural damage. 
% While it is possible for every robot to be a leader (have access to desired trajectory), it suffers from the need of sufficient communication bandwidth \cite{Montemayor2005}.
If all the robots are leaders with access to the desired transport trajectory, then the network response would be cohesive, but this  leads to a centralized approach, and such communication might not be always feasible, e.g., if one of the robots is directly controlled by a human and the others follow based on neighbor-based observations or local sensing of the object. This motivates the current effort to improve cohesion during decentralized transport of flexible objects.

\vspace{0.1in}
This work aims to reduce deformations of the object being transported by developing methods for cohesive positioning of the robot network.
Recent studies have shown that control laws can be developed to improve cohesion of decentralized networks using Delayed Self Reinforcement (DSR)~\cite{Devasia2020}.  The main contribution of the current paper is to develop an approach to transport objects using the cohesive method in~\cite{Devasia2020}, without the need for inter-robot communications - rather, only the local force measurements are required. Specifically, the current work  (i)~shows that the DSR approach can be used to achieve cohesive transport using only local force measurements,  without the need of inter-robot communication, and (ii)~establishes stability conditions for the discrete-time  implementation of the proposed cohesive transport approach. 
Furthermore,  experimental results are used to show that the proposed approach improves network cohesion, and leads to low-deformation transport of flexible objects.

\newpage
\section{Problem formulation}
In this section, the deformation control issue with local force-based decentralized transport dynamics is presented along a single axis-$y$. Similar approaches can be used for the other axes of motion. 
% An example of the single axis transport network is shown in Figure~\ref{experiment_schematic}. 

\vspace{-0.05in} 
\subsection{Local-force feedback as a network-based update}
\begin{figure}[ht!]
  \centering
  \includegraphics[width=0.95\columnwidth]{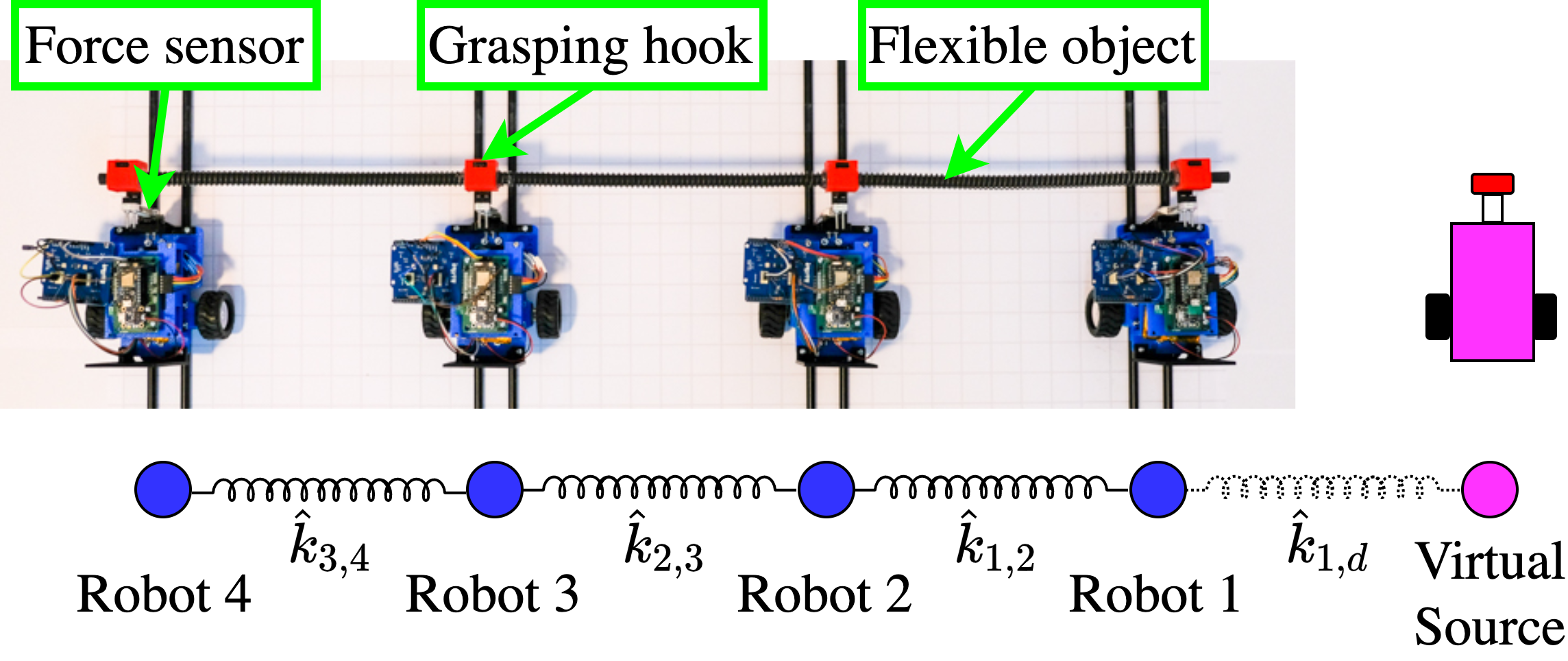}
  \caption{Top: Experimental setup of flexible load transport. Bottom: Schematic network model where  the leader, robot $k=1$, aims to match the position of the virtual source (pink).}
  \label{experiment_schematic}
\end{figure}

The robots are attached to a flexible object, e.g., as shown in Fig.~\ref{experiment_schematic}. The position 
 $y_k \in \mathbb{R}$ of each robot $k$ in the network is updated using local force measurements $f_k \in \mathbb{R}$ as well as a virtual force $\tilde{f}_k =  \hat{k}_{k,d} (y_k - y_d) \in \mathbb{R}$ if robot $k$ is a leader, as 
    \begin{align}
         y_k[m+1] &  = y_k[m] - \gamma \hat{f}_k[m],  \label{Eq_single_dynamics}  \\
        \hat{f}_k[m] & = f_k[m] + \tilde{f}_k[m],  
        %\nonumber \\
       % & = f_k[m] + \hat{k}_{k,d} (y_k[m] - y_d[m]),
           \label{Eq_force_with_source}
    \end{align}
where the update sampling-time period is $\delta_t$, $y_k[m]$ represents the position of robot $k$ at discrete time instants, e.g., $y_k[m] = y_k(m\delta_t)$, $\gamma$ is the update gain, and  the desired position from the virtual source $y_d \in \mathbb{R}$ is known (i.e., 
$ \hat{k}_{k,d} \ne 0$) only if robot $k$ is a leader, e.g., $k=1$ in the example in Fig.~\ref{experiment_schematic}. Each robot position  $y_k$ is measured from an initial undeformed configuration (with all $y_k=0$) of the flexible object. 

The  local force-based robot position update in Eq.~\eqref{Eq_single_dynamics} achieves object transport, i.e., for a fixed desired position $y_d$, the network reaches equilibrium  when each robot reaches the desired position, i.e.,  $y_k = y_d$ for all $k$ and the object has no distortion, i.e., $f_k = 0$. To clarify, note that  the local force $f_k$ measured by robot $k$ can be written linearly in terms of the position $y_j$ of all robots connected to robot $k$,
provided the local deformation of the flexible object remains small, as 
    \begin{equation}
        f_k[m] = \sum_{j=1}^{n} \hat{k}_{k,j} (y_k[m] - y_j[m]),
        \label{Eq_force_single}
    \end{equation}
where $\hat{k}_{k,j} = \hat{k}_{j,k} \geq 0 $ is the effective stiffness of the flexible object between two neighboring robots $j$ and $k$, $n$ is the number of robots, and $\hat{k}_{j,j} =0$ for all  $1\le j \le n$. Then, the  update law in Eq.~\eqref{Eq_single_dynamics} can be written in matrix form as
    \begin{align}
        \mathbf{Y}[m+1] & = (\mathbf{I} - \gamma \mathbf{K}) \mathbf{Y}[m] + \gamma \mathbf{B} y_d[m], 
            \label{Eq_matrix_dynamics}
    \end{align}
where $\mathbf{Y} = [y_1, y_2, \hdots , y_k, \hdots,  y_n]^T \in \mathbb{R}^n$ is the $n$ dimensional vector of the individual robots positions $y_k$, 
$B_k = \hat{k}_{k,d}$ is the $k^{th}$ element of $n$ dimensional vector $\mathbf{B} \in \mathbb{R}^n$ which is nonzero only if the robot $k$ is a leader 
and the symmetric matrix $\mathbf{K}$ is the pinned Laplacian with elements
\begin{equation}
    \begin{array}{rcll}
         {\mathbf{K}}_{k,j} & = & \hat{k}_{k,d} + 
         \sum_{m=1}^{n } \hat{k}_{k,m} ~~ \ge 0  & 
    {\mbox{if}}~ k =j
        \\[0.3em] 
       %\hat{k}_{i,m} 
        & = &  - \hat{k}_{k,j} &  {\mbox{otherwise}}, 
    \end{array}
        \label{force_eq_Sd_3}
\end{equation}
with real nonzero eigenvalues $\lambda_{\mathbf{K},j}>0$ for $1 \le j \le n$.
It can be shown that for an update gain $\gamma$ satisfying  
\begin{equation}
        0 < \gamma < \left( \overline{\gamma} ~= 2/\overline{\lambda}_{\mathbf{K}}  ~= 2/ \max_j ( \lambda_{\mathbf{K},j} ) 
       % \frac{2}{\overline{\lambda_K}}  
        \right),
            \label{Eq_stability_bound_gamma}
\end{equation}
the eigenvalues of $(\mathbf{I} - \gamma \mathbf{K})$ are inside the unit circle. 
Therefore, the transport dynamics in Eq.~\eqref{Eq_matrix_dynamics} is stable, and the desired transport of the object can be achieved, i.e.,  for a fixed desired position $y_d$, the robot positions in  $\mathbf{Y}$  converge to the desired value~\cite{Olfati_Murray_07}
\begin{equation}
\lim_{m \rightarrow \infty} \mathbf{Y}[m] ~= \textbf{1}_n y_d  
\end{equation}
with 
$\textbf{1}_n$ representing the $n$ dimensional vector of all ones.

\subsection{Problem: improve cohesion for similar settling time}
The settling time $T_s$ to a new desired position $y_d$ can be selected by choosing the update gain $\gamma$. In particular, 
the settling time $T_s$ to reach and stay within 2\% of a step change in the desired displacement $y_d$ can be estimated as 
\begin{equation}
    T_s \approx \frac{-4~\delta_t}{\ln\left( \lambda_\mathbf{K}^* \right)}   ~~{\mbox{where}} 
    ~ \lambda_\mathbf{K}^* = 
    \argmax_{\lambda_{\mathbf{K},j}}{|1 - \gamma \lambda_{\mathbf{K},j} |} .
    \label{Eq_settling_time}
\end{equation}
However, for a given settling time (i.e., given selection of the update gain $\gamma$), the deformations during transport can not be controlled further. Typically, faster settling (i.e., a smaller settling time $T_s$) also results in larger deformation. The research problem addressed here is to improve cohesion, i.e., to reduce the maximum deformation $\overline{D}$ of the object during transport, 
\begin{equation}
    \overline{D} = \max_m  \left[ D[m] = \max_{k,j} \big|\:  (y_j[m] - y_k[m]) \: \big| \right],
    \label{Eq_max_deformation}
\end{equation}
without increasing the settling time $T_s$.
\vspace{0.1in}
\section{Transport using cohesive DSR}
\subsection{Cohesive transport using local force measurements}
The robot-position update to transport the object is chosen as a discrete-time approximation of an ideal cohesive network. 
For example, in the continuous time case, if each robot had access to the desired position $y_d$, i.e., in a centralized approach, then the ideal cohesive update can be written as, 
\begin{equation}
    \dot{\mathbf{Y}}(t) = -\alpha \mathbf{Y}(t) + \alpha \textbf{1}_n y_d(t), 
    \label{Eq_ideal_motion}
\end{equation}
where the gain $\alpha > 0$ can be tuned to achieve a desired  settling time and all robot move in a similar manner. 
The lack of access to centralized information about the desired position $y_d$ can be alleviated by 
multiplying both sides with the scaled pinned Laplacian $\beta \mathbf{K}$, substituting $\mathbf{K}\textbf{1}_n$ with $\mathbf{B}$ ~\cite{Devasia2020}, and adding $\dot{\mathbf{Y}}$ to both sides and rearranging the equation to obtain
\begin{align}
    \dot{\mathbf{Y}}(t) & = -\alpha \beta \mathbf{KY}(t) +  \alpha \beta \mathbf{B} y_d(t) + [\mathbf{I} - \beta \mathbf{K}]\dot{\mathbf{Y}}(t),
\label{eq_continuous_time_ideal} \\
& \approx 
 -\alpha \beta \mathbf{KY}(t) +  \alpha \beta \mathbf{B} y_d(t) + \left[\mathbf{I} - \beta \mathbf{K} \right] \frac{\mathbf{Y}(t) - \mathbf{Y}(t - \tau)}{\tau}.
      \label{eq_cohesive_dynamics}
\end{align}
With the time delay $\tau = \delta_t$
and the update kept constant between sampling periods, 
the update law in Eq.~\eqref{eq_cohesive_dynamics} for robot positions becomes 
\begin{equation}
    \begin{aligned}
        \mathbf{Y}[m+1] = & \mathbf{Y}[m] -\alpha \beta \delta_t \mathbf{KY}[m] +  \alpha \beta \delta_t \mathbf{B} y_d[m] \\
        & + \left[\mathbf{I} - \beta \mathbf{K} \right] \left(\mathbf{Y}[m] - \mathbf{Y}[m - 1]\right).
    \end{aligned}
    \label{Eq_discrete_matrix}
\end{equation}
Then for each robot $k$, the update law becomes  
    \begin{align}
        y_k[m+1] = & y_k[m] -\alpha \beta \delta_t f_k[m] + \alpha \beta \delta_t \hat{k}_{k,d} \big(  y_d[m] - y_k[m] \big) \nonumber \\
        & {+ (1-\beta\hat{k}_{k,d}) \big( y_k[m] - y_k[m-1] \big)}  \nonumber \\
        & - \beta \big( f_k[m] - f_k[m-1] \big).
        \label{Eq_local_force_implementation}
    \end{align}
Note that this cohesive transport law is decentralized. For each robot, in addition to
the virtual source position $y_d$ if the robot is a leader, 
the update only requires current  information and delayed self reinforcement (DSR) by a time step of: (i) local force measurements $f_k$ and (ii) its own position $y_k$.

\vspace{0.1in}

\subsection{Stability of cohesive DSR}
Stability conditions can be established by finding the roots of the characteristic equation of the cohesive dynamics in Eq.~\eqref{Eq_discrete_matrix}, i.e., 
\begin{equation}
\begin{aligned}
    &\det \Big( 
        \mathbf{I} z^{2}   -  
        \left(
        \mathbf{I}-\alpha \beta \delta_t \mathbf{K}_J + [\mathbf{I}-\beta \mathbf{K}_J]  \right)z 
        +[\mathbf{I}-\beta \mathbf{K}_J ] \Big) = 0,
\end{aligned}
\label{Eq_characteristic_Eq}
\end{equation}
where  $\mathbf{K}_J  = \mathbf{P_K}^{-1} \mathbf{K P_K} $ is the diagonalization of the pinned Laplacian $\mathbf{K}$ with eigenvalues $\lambda_{\mathbf{K},k}$ along the diagonal. In particular,  
the cohesive dynamics in Eq.~\eqref{Eq_discrete_matrix} is stable 
if and only if, for each eigenvalue $\lambda_{\mathbf{K},k}$ of the pinned Laplacian $\mathbf{K}$, the roots of $D(z)$ where 
\begin{equation}
\begin{aligned}
    D(z) = & z^{2} - \left(1-\alpha \beta \delta_t \lambda_{\mathbf{K},k} + [1-\beta \lambda_{\mathbf{K},k}]\right)z
    \\[0.3em]
    & + [1-\beta \lambda_{\mathbf{K},k}]  = 0
    \end{aligned}
    \label{Eq_characteristic}
\end{equation}
have magnitude less than one. 
Thus, stability can be evaluated by computing the roots of Eq.~\eqref{Eq_characteristic}. Nevertheless, for design purposes, it is preferable to establish analytical conditions on the DSR parameters for stability, as shown next.

\vspace{0.1in}

\begin{lemma}
\label{Lemma_N_1}
The proposed cohesive DSR based dynamics in Eq.~\eqref{Eq_discrete_matrix} is stable if and only if the gains $\alpha, \beta$ satisfy the following conditions for the largest eigenvalue $\overline{\lambda}_{\mathbf{K}}$ of the pinned Laplacian $\mathbf{K}$:
\begin{equation}
    \begin{aligned}
        (i) &\quad  0 < \alpha \\
        (ii) &\quad  0 < \beta < \frac{4}{\overline{\lambda}_{\mathbf{K}}(\alpha \delta_t  + 2 )}.
    \end{aligned}
    \label{Eq_lemma_1}
\end{equation}
\end{lemma}

\vspace{0.1in}

\noindent \textit{Proof:} Proof of Lemma \ref{Lemma_N_1} follows from Jury stability test and is omitted here for brevity.

\vspace{0.1in}

\begin{Rem}
The delay in Eq.~\eqref{eq_cohesive_dynamics} can be defined over multiple samples, i.e., $\tau = N \delta_t$ where $N \geq 1$ is an integer number. 
However a larger delay $\tau$ (larger $N$) reduces the effectiveness of the approximate derivative in Eq.~\eqref{eq_cohesive_dynamics} and thereby, reduces the ability to track faster signals in a cohesive manner~\cite{Devasia2020}. 
\end{Rem}

\vspace{0.1in}

\section{Experiment and parameter selection}
\subsection{System description} 
\begin{figure}[!ht]
  \centering
  \includegraphics[width=0.95\columnwidth]{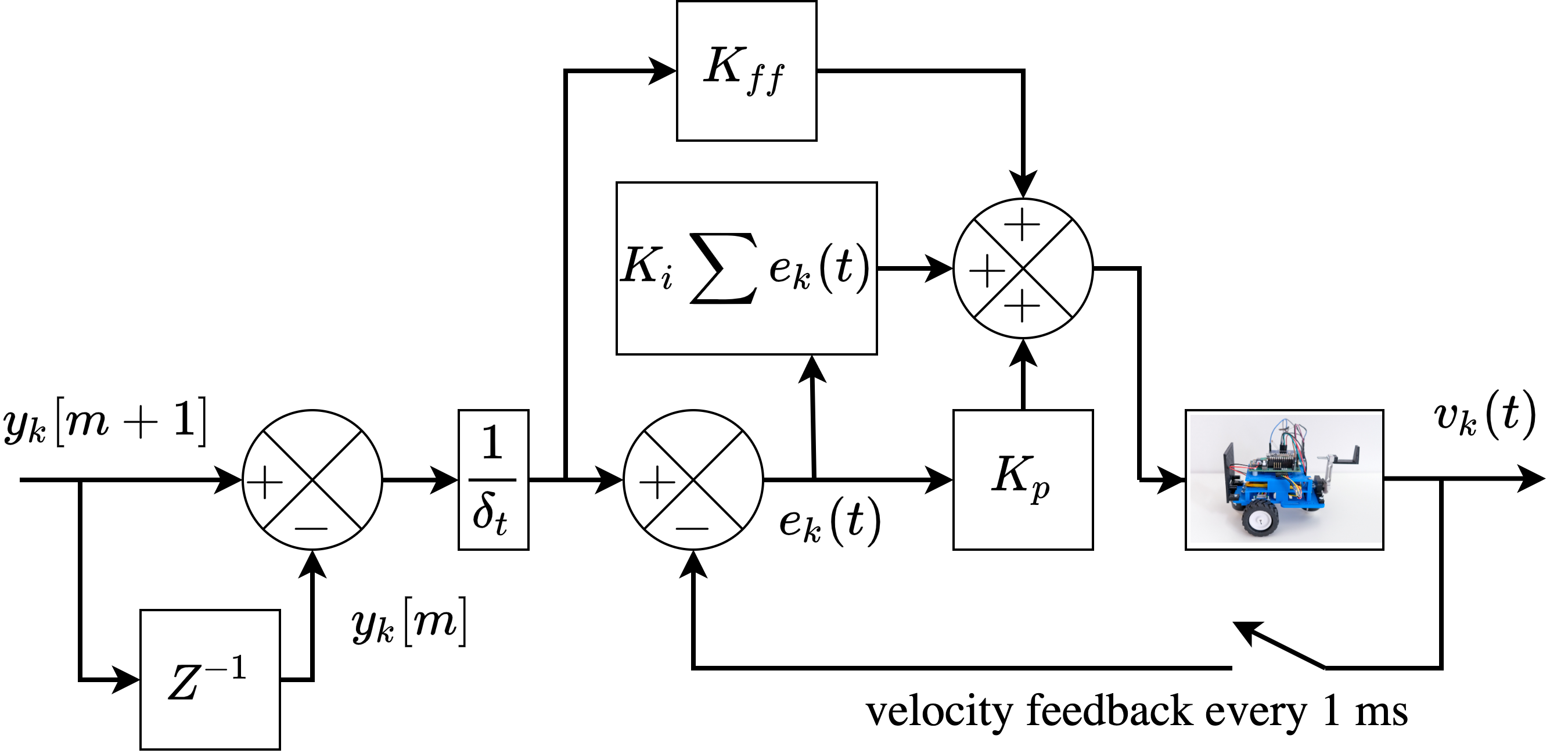}
  \caption{Velocity-based feedback control system of  robot $k$ to achieve the position $y_k[m+1]$ using a proportional (gain $K_p$) and integral (gain $K_i$) feedback controller along with feedforward 
  control (gain $K_{ff}$).
  }
  \label{bot_controller}
  \vspace{0.1in}
\end{figure}

To easily visualize the deformation during transport, a highly flexible object (a long spring coiled with diameter of $1.30\;cm$ and length of $90\;cm$) was selected for transport using a robot network as shown in Fig.~\ref{experiment_schematic}. Only the leader (robot 1) has knowledge of the desired position illustrated by the virtual source shown in pink in Fig.~\ref{experiment_schematic}.
Each robot $k$ measures the local force $f_k$ using a force sensor between the robot and the object, and senses its position $y_k$ using magnetic encoders on the wheels, and uses an on-board micro-controller to compute the next position $y_k[m+1]$. The updated position $y_k[m+1]$ is achieved within the sampling-time period $\delta_t=0.03\;s$, using a velocity feedback control to maintain the velocity at $v_{d,k}(t) = (y_k[m+1] - y_k[m])/\delta_t$ as shown in Fig.~\ref{bot_controller}.

The system dynamics in Eq.~\eqref{Eq_single_dynamics} was found by estimating the elastic object stiffness
$\hat{k}_{i,j}$ experimentally. In particular, 
for estimating $\hat{k}_{1,2}$, robot 1 was moved for a known distance $y_1$ without moving the other robots (nor connecting robot 1 to the virtual source) and the resulting force $f_1$ was measured to yield $\hat{k}_{1,2} = f_1/y_1$. Then, to estimate stiffness $\hat{k}_{2,3}$, robot 2 was moved for a known distance $y_2$ without moving the other robots and the resulting force $f_2$ was measured to yield 
$\hat{k}_{2,3} = (f_2/y_2) - \hat{k}_{1,2}$.  The same procedure is used to obtain the rest of the  effective stiffness coefficients, which are all the same and given by 
$\hat{k}_{i,i+1} = 0.05\;N/cm, \; {\mbox{for}} \;  1\le i \le 3, $ which is to be expected since  the connecting springs have similar lengths. For the setup shown in Fig.~\ref{experiment_schematic}, the stiffness of the connection with virtual source is chosen to be same as the other spring elements, i.e., $\hat{k}_{1,d} = 0.05$. The resulting  pinned Laplacian $\mathbf{K}$ and matrix $\mathbf{B}$ were %
\begin{align}
    \mathbf{K} & =  
    \begin{bmatrix}
    \hat{k}_{1,2} + \hat{k}_{1,d} & -\hat{k}_{1,2} & 0 & 0\\
    -\hat{k}_{1,2} & \hat{k}_{1,2} + \hat{k}_{2,3} & -\hat{k}_{2,3} & 0 \\
    0 & -\hat{k}_{2,3} & \hat{k}_{2,3} + \hat{k}_{3,4} & -\hat{k}_{3,4} \\
    0 & 0 & -\hat{k}_{3,4} & \hat{k}_{3,4}
    \end{bmatrix} \nonumber \\[0.3em]
    & = 
    \begin{bmatrix}
    0.10 & -0.05 & 0 & 0\\
    -0.05 & 0.10 & -0.05 & 0 \\
    0 & -0.05 & 0.10 & 0.05 \\
    0 & 0 & -0.05 & 0.05
    \end{bmatrix}, 
        \label{dynamics_eq_Sd_4_example}
% \end{align}
\\[0.3em]
    %
% \begin{align}
    \mathbf{B} & = 
    \begin{bmatrix}
    \hat{k}_{1,d} & 0 & 0 & 0 
    \end{bmatrix}^T ~= 
    \begin{bmatrix}
    0.05 & 0 & 0 & 0  
    \end{bmatrix}^T.  %\nonumber 
\end{align}
\vspace{0.1in}
\subsection{Selection of control parameters}
To avoid optimization over each desired trajectory $y_d$, the control parameters are selected to minimize the deformation for a specified settling time $T_s$ for a unit step change in the desired position $y_d$. 

\vspace{0.1in}

\subsubsection{Case without DSR}
The update gain $\gamma$ is found numerically for a specified  network settling time $T_s$ when the position changes by a unit step. 
For update gains $\gamma$  satisfying the stability condition in Eq.~\eqref{Eq_stability_bound_gamma}, the settling times $T_s$ were estimated using Eq.~\eqref{Eq_settling_time}, and are shown in
Fig.~\ref{Ts_vs_gamma}. 
Interpolation of this data can be used to find the update gain $\gamma$ for a specified settling time $T_s$. In the following, the settling time $T_s = 10\;s$ is chosen in order to bound the maximum speed input to the robot $v_{d,k} = v_{nodsr}$ below the acceptable speed limit $v_{max} = 5\;cm/s$ as shown in Fig.~\ref{max_vel}. The corresponding update gain $\gamma = 1.93$ and the step response is shown in Fig.~\ref{step_traj}.

\vspace{0.1in}

\subsubsection{Case with cohesive DSR}
\begin{figure}[ht!]
  \centering
  \subfloat[ ]
  {
    \includegraphics[width=0.47\columnwidth]{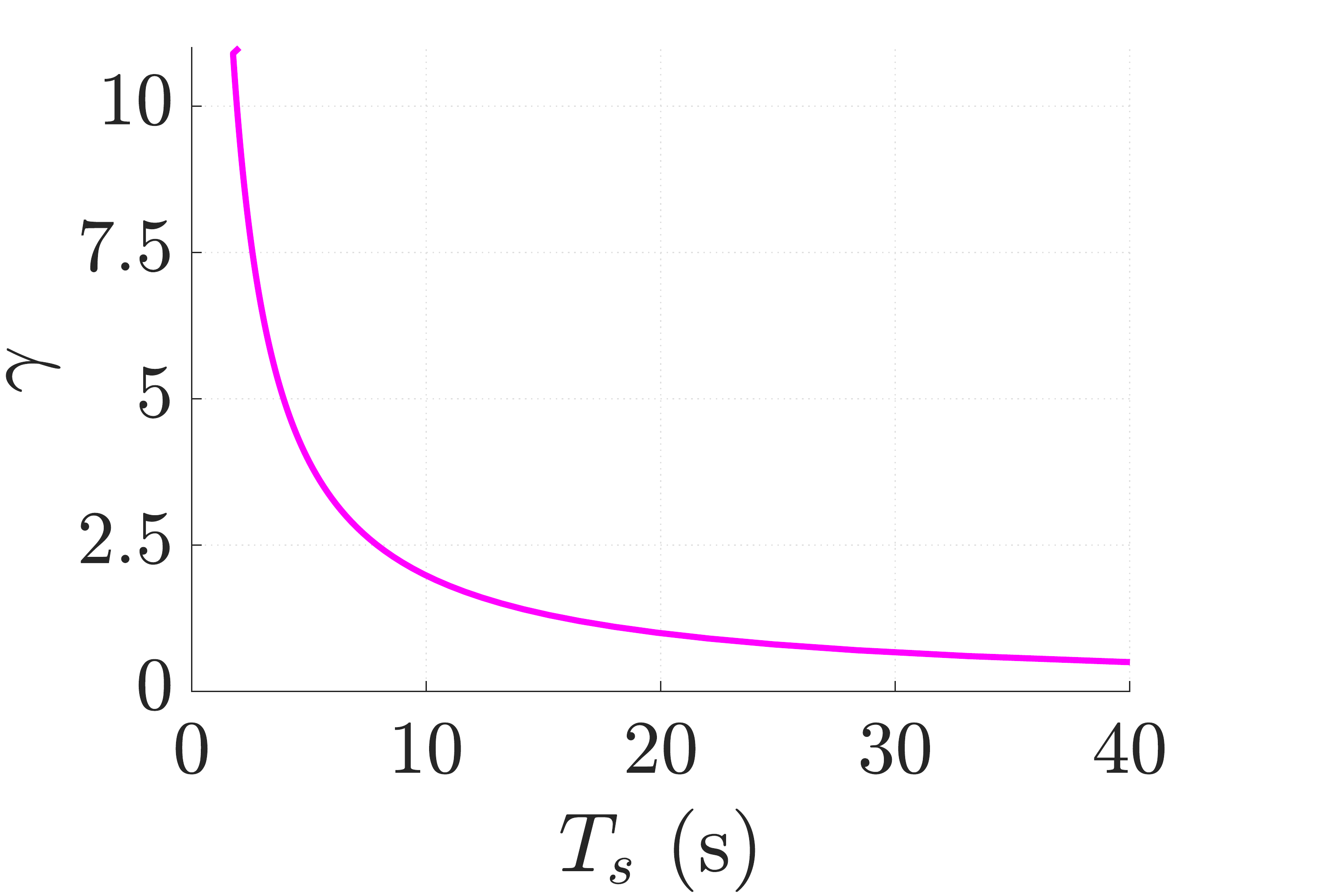}
    \label{Ts_vs_gamma}
  }
  \subfloat[ ]
  {
    \includegraphics[width=0.47\columnwidth]{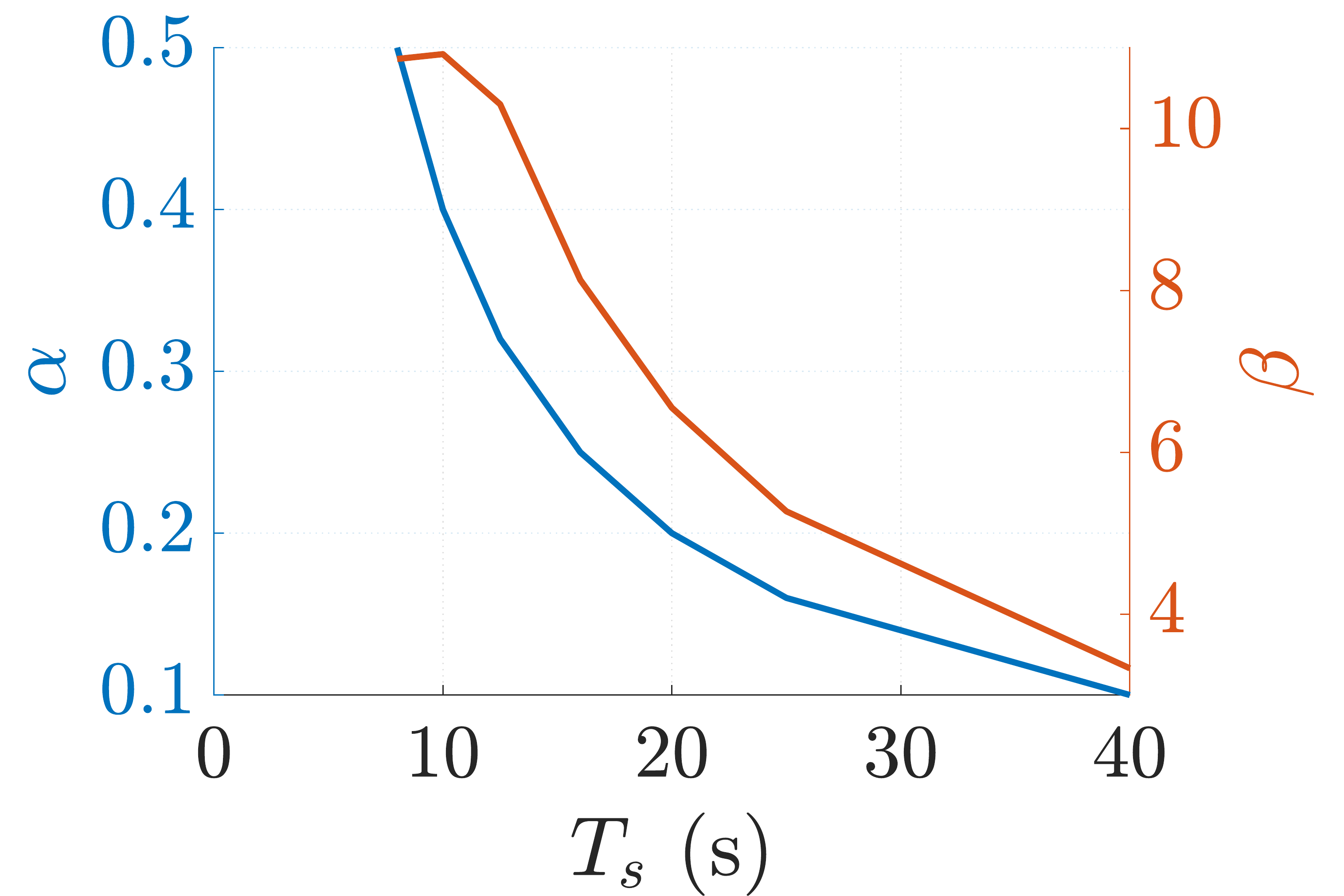}
    \label{Ts_vs_dsrParam}
  }
  
    \caption{Selection of control parameters with respect to settling time $T_s$: (left) The update gain $\gamma$ for the case without DSR, (right) DSR parameters $\alpha \mbox{ and } \beta$ for the case with cohesive DSR. 
    }
  \label{Ts_vs_parameters}
\end{figure}

\begin{figure}[ht!]
  \centering
  \subfloat[ ]
  {
    \includegraphics[width=0.68\columnwidth]{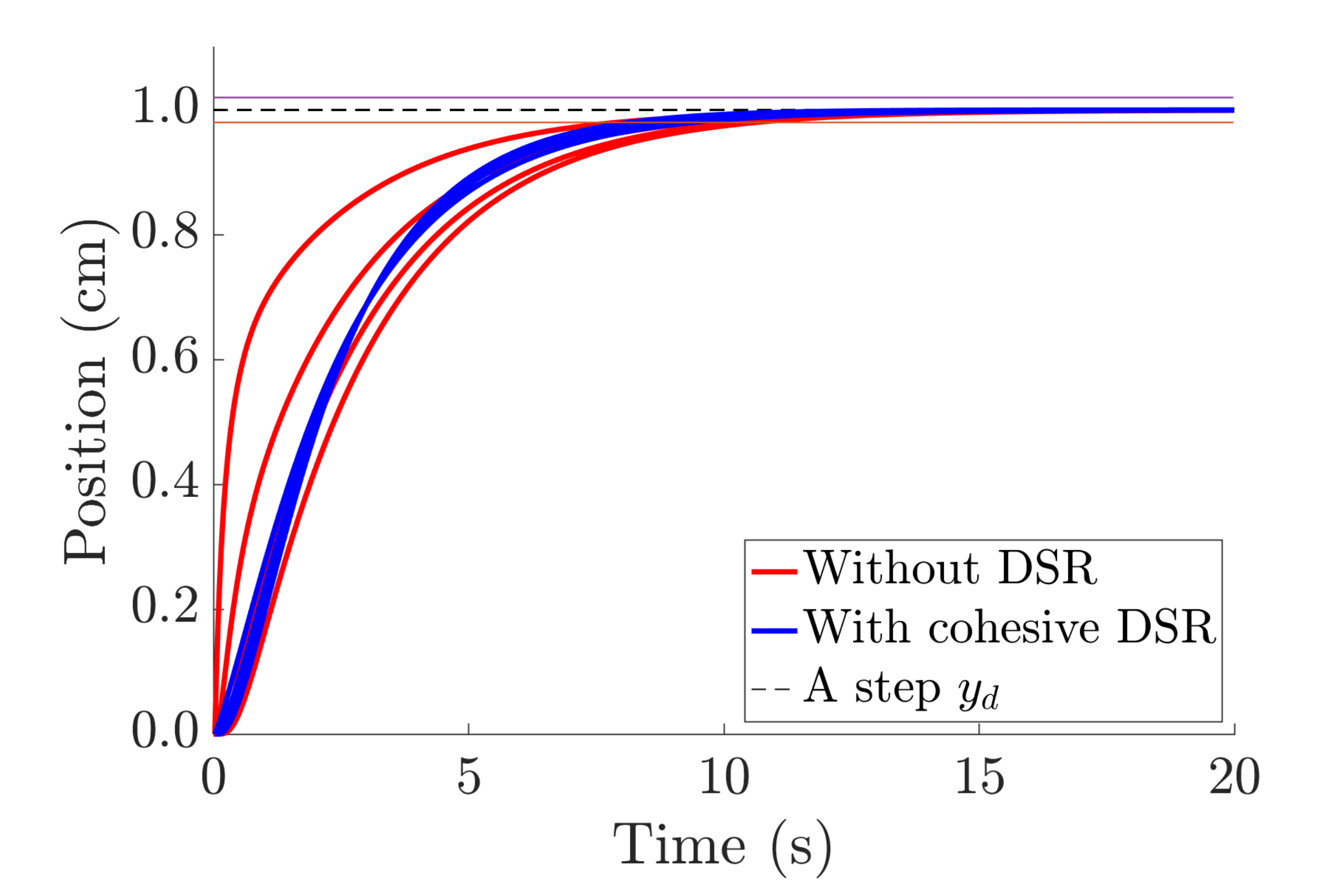}
    \label{step_traj}
  }
  \\
  \subfloat[ ]
  {
    \includegraphics[width=0.68\columnwidth]{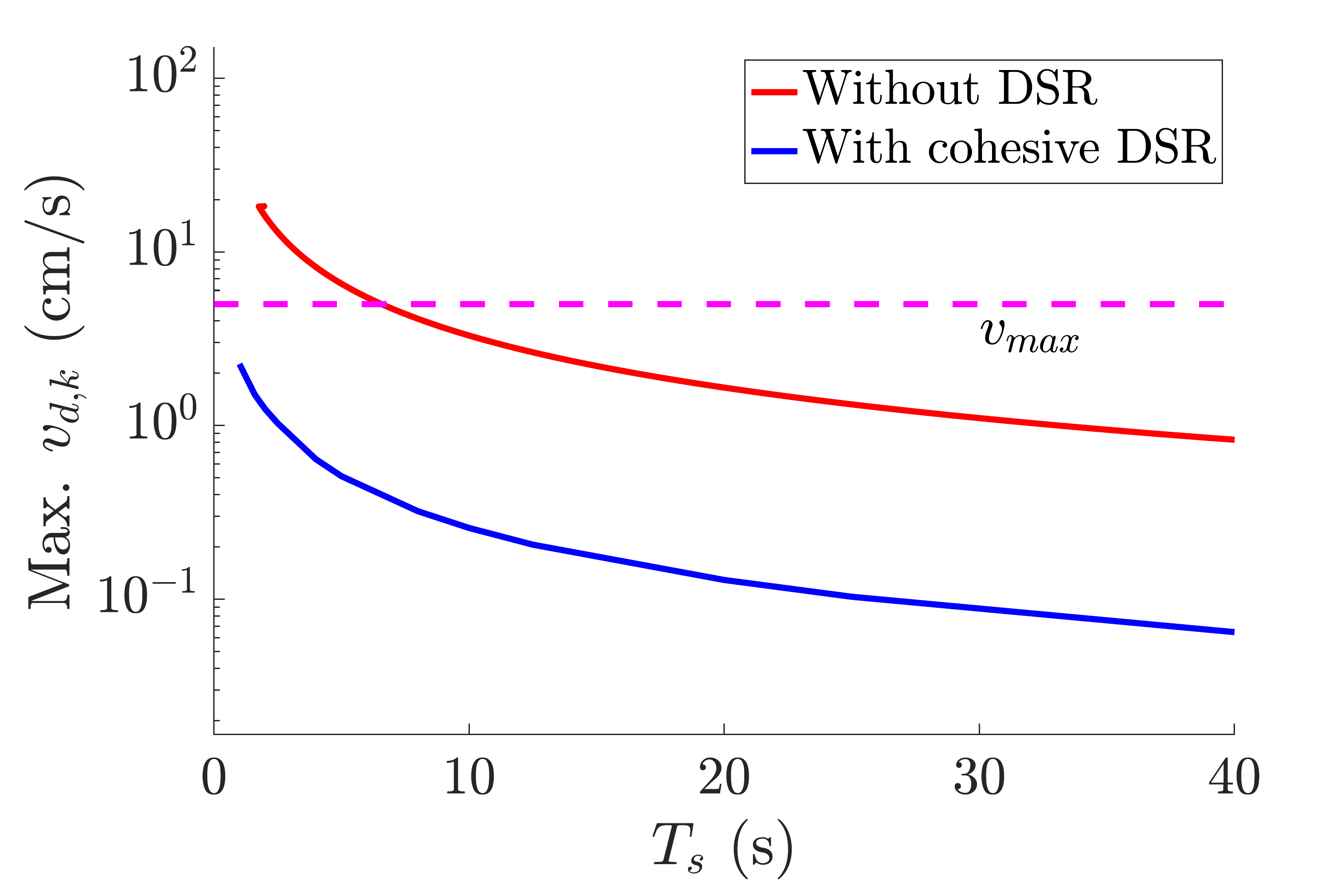}
    \label{max_vel}
  }
  
    \caption{ 
    {\bf{(a) Increase in cohesion with DSR.}}  Position responses for a step change in position $y_d$ for both (i)~without DSR ($\gamma = 1.93$) as in Eq. ~\eqref{Eq_matrix_dynamics} and (ii)~with cohesive DSR ($\alpha = 0.39$, $\beta = 10.92$) as in Eq.~\eqref{Eq_discrete_matrix} which show that both settle within $T_s = 10\;s$.  {\bf{ (b) Reduction of maximum input speed with DSR for the same settling time.}} 
    Selection of settling time $T_s$ such that the maximum speed $v_{d,k}$ of the robot is below the acceptable speed limit $v_{max}$.
    % {\color{red} \hl{Yoshua: use red for no DSR and blue for DSR case  to match the plot colors for both cases. green and blue are in general hard to distinguish!  }}
    }
  \label{step_pos_and_vel}
%   \vspace{-0.1in}
\end{figure}

To enable comparative evaluation, the cohesive DSR parameters ($\alpha,\beta$) are selected to match the settling time $T_s$ of the case without DSR, and the maximum speed input $v_{d,k}$ is below $v_{nodsr}$ for the case without DSR. Since it is possible to obtain multiple combinations of the parameters ($\alpha,\beta$) that satisfy the settling time $T_s$ and the maximum speed input $v_{d,k}$ conditions,  the optimal parameters are  selected such that the spectral radius $\sigma$ is minimized (to maximize structural robustness), 
i.e., 
    \begin{equation}
    \begin{aligned}
         \sigma^{*} = &\min_{\alpha, \beta} \left(\sigma(\alpha, \beta) =  \max_j |z_{j}| \right), \\
        \; &\mbox{subject to} \; v_{d,k} \leq v_{nodsr}
        \end{aligned}
        \label{Eq_spec_radius_DSR}
    \end{equation}
where $z_j$ is the $j^{th}$ root of the characteristic equation $D(z)$ as in Eq.~\eqref{Eq_characteristic}. 
The parameters ($\alpha,\beta$) for the same range of settling time $T_s$ as in the case without DSR are shown in
Fig.~\ref{Ts_vs_dsrParam}. In the following, the settling time $T_s = 10\;s$ is chosen to match the case without DSR and the corresponding parameters are
    \begin{equation}
         \alpha = 0.39~ \approx 0.4 , \qquad \beta = 10.92~ \approx 10.9.
         \label{eq_numerical_alpha_beta}
    \end{equation}
The step response is also shown in Fig.~\ref{step_traj}. Note that the maximum speed input to the robot $v_{d,k}$ is also well below $v_{nodsr}$ as shown in Fig.~\ref{max_vel}.

\begin{Rem}
When the sampling time $\delta_t$ becomes small as compared to the transport time, the discrete time cohesive dynamics in Eq.~\eqref{Eq_discrete_matrix} should  be similar to the continuous-time ideal cohesive dynamics in Eq.~\eqref{Eq_ideal_motion}. Therefore, the settling time becomes $T_s = 4/\alpha$ and can be selected using the parameter $\alpha$. With the settling time chosen as $T_s = 10\;s$, the estimate of $\alpha = 4/T_s = 0.4$ is close to the result from the numerical search $\alpha = 0.39 $ in Eq.~\eqref{eq_numerical_alpha_beta}. 
\end{Rem}
\vspace{0.1in}

\begin{Rem}
The spectral radius, provided the associated second-order dynamics in Eq.~\eqref{Eq_characteristic} is not overdamped, is the maximum value of $| 1 - \beta \lambda_{\mathbf{K},k}| $, which is minimized over all eigenvalues $\lambda_{\mathbf{K},k}$, by selecting $ \beta = \frac{2}{\underline{\lambda}_\mathbf{K} + \overline{\lambda}_\mathbf{K}} =  \frac{2}{0.006 + 0.176}~ =  10.95$, which is close to the result from the numerical search,  $\beta = 10.92$ in Eq.~\eqref{eq_numerical_alpha_beta}.
\end{Rem}
\begin{figure*}[!ht]
  \centering
  \subfloat[Without DSR\label{fig_Force_Plot_without_DSR_sim}]
  {
    \includegraphics[width=0.65\columnwidth]{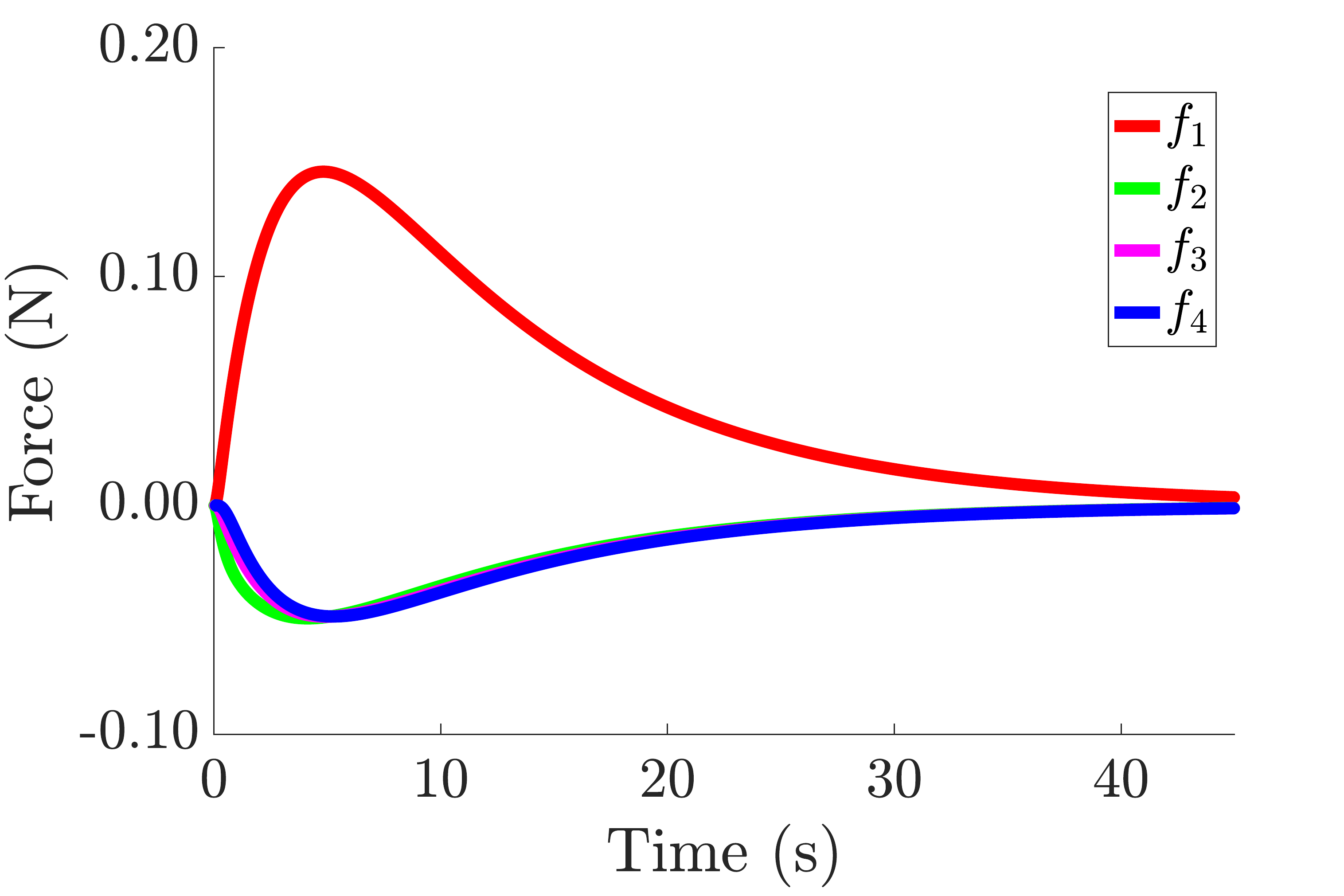}
  }
  \subfloat[With cohesive DSR\label{fig_Force_Plot_with_DSR_sim}]
  {
    \includegraphics[width=0.65\columnwidth]{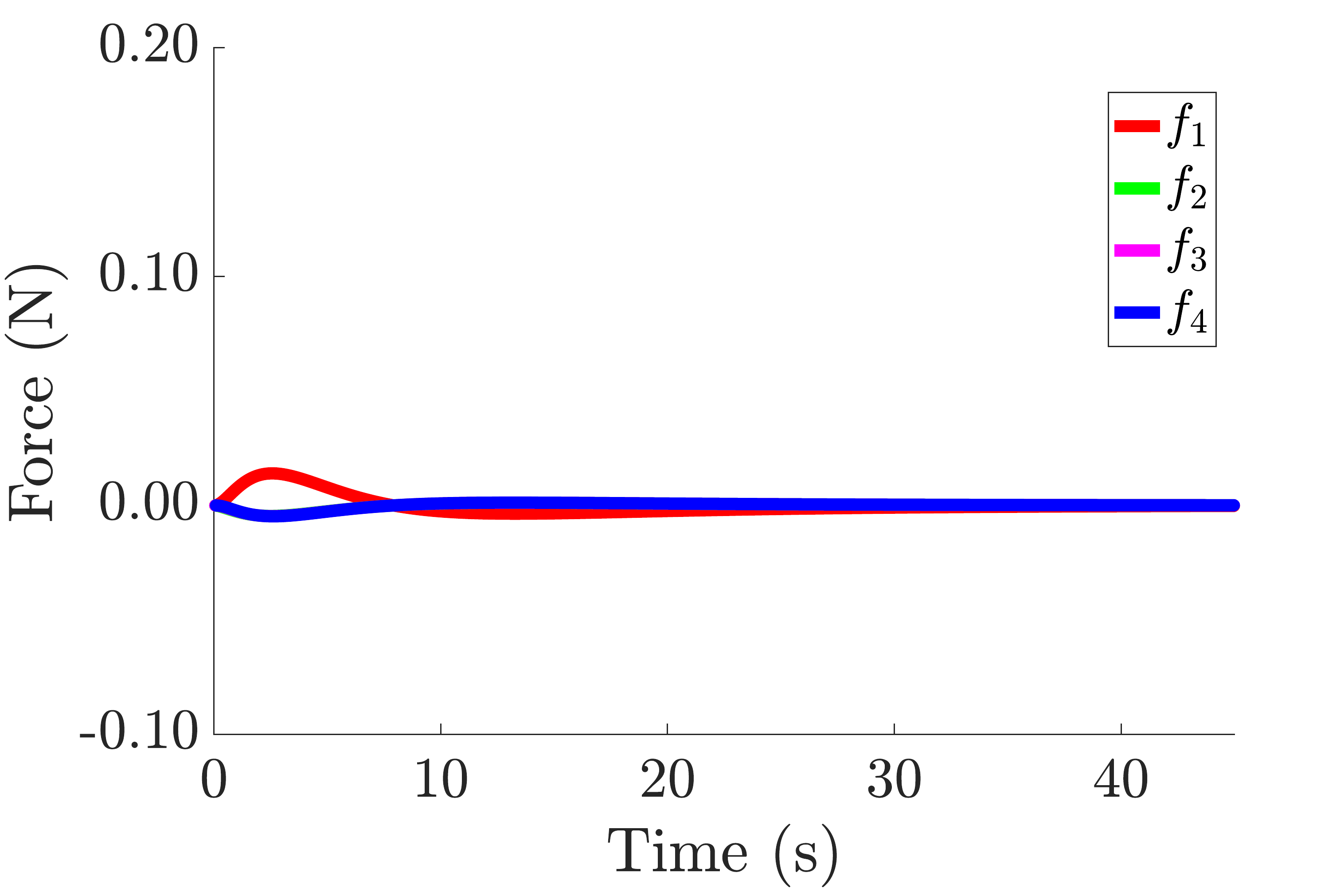}
  }
\subfloat[Deformations\label{fig_Deform_Plot_sim}]
  {
    \includegraphics[width=0.65\columnwidth]{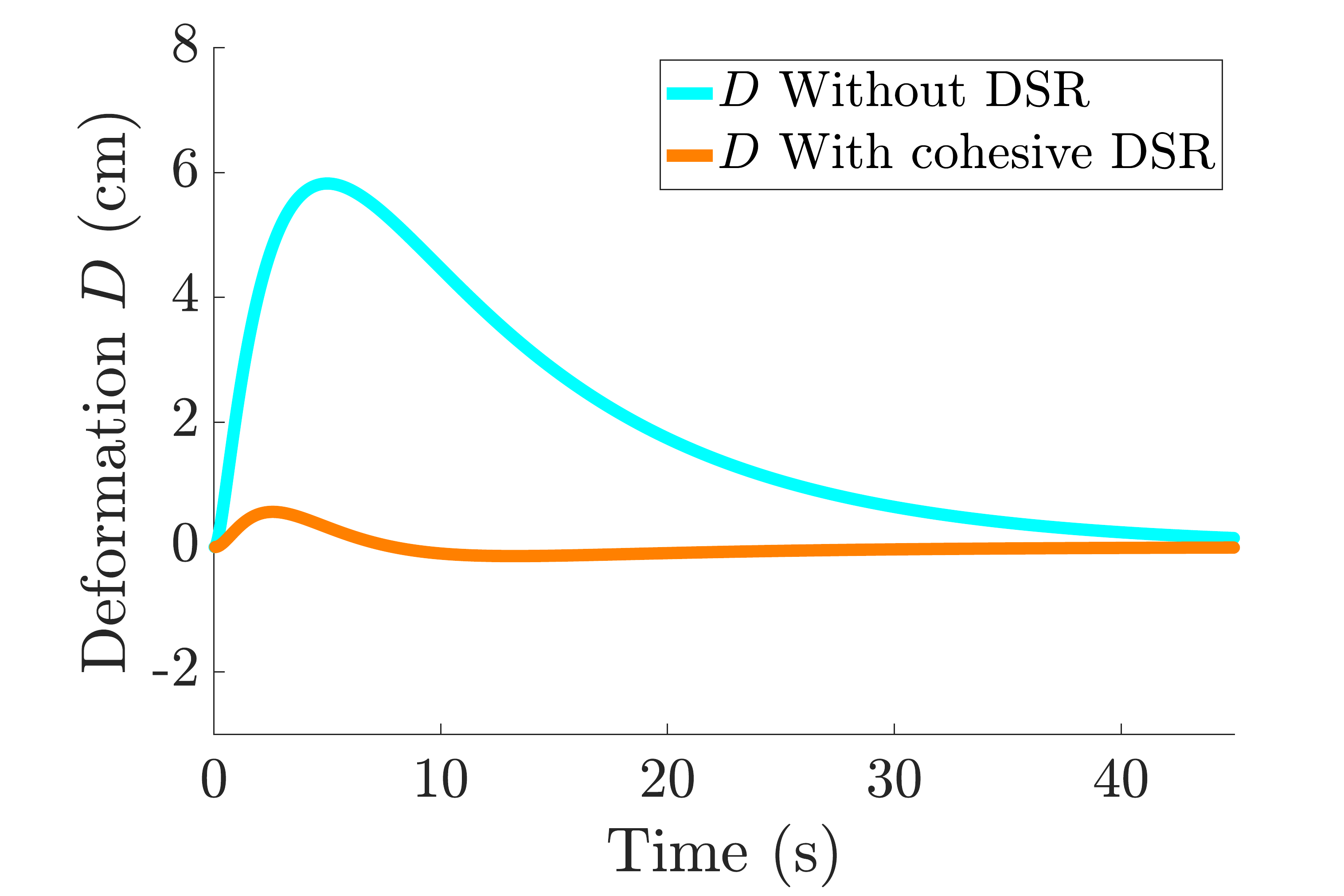}
  }\\
    \subfloat[Without DSR\label{fig_Force_Plot_without_DSR_exp}]
  {
    \includegraphics[width=0.65\columnwidth]{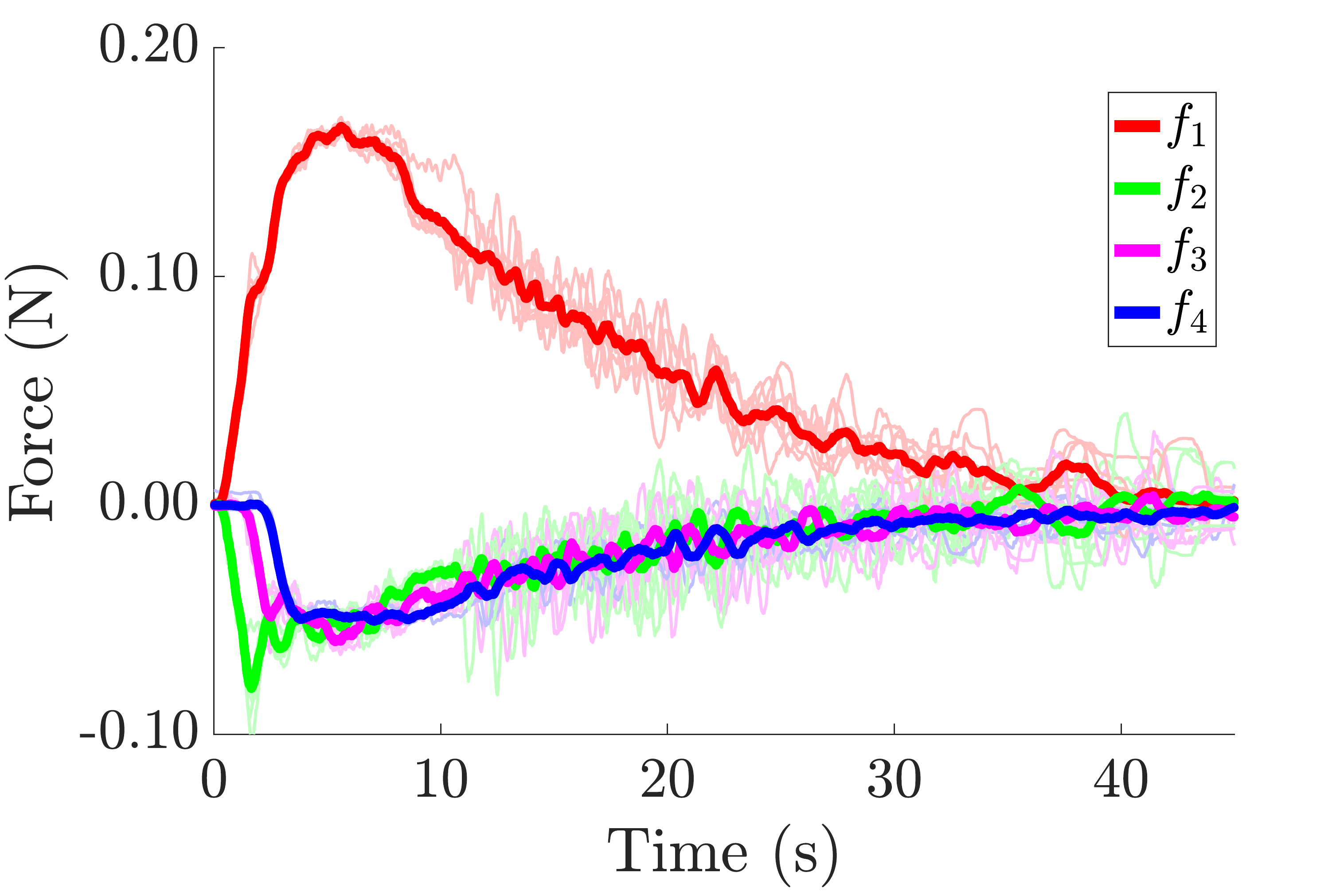}
  }
  \subfloat[With cohesive DSR\label{fig_Force_Plot_with_DSR_exp}]
  {
    \includegraphics[width=0.65\columnwidth]{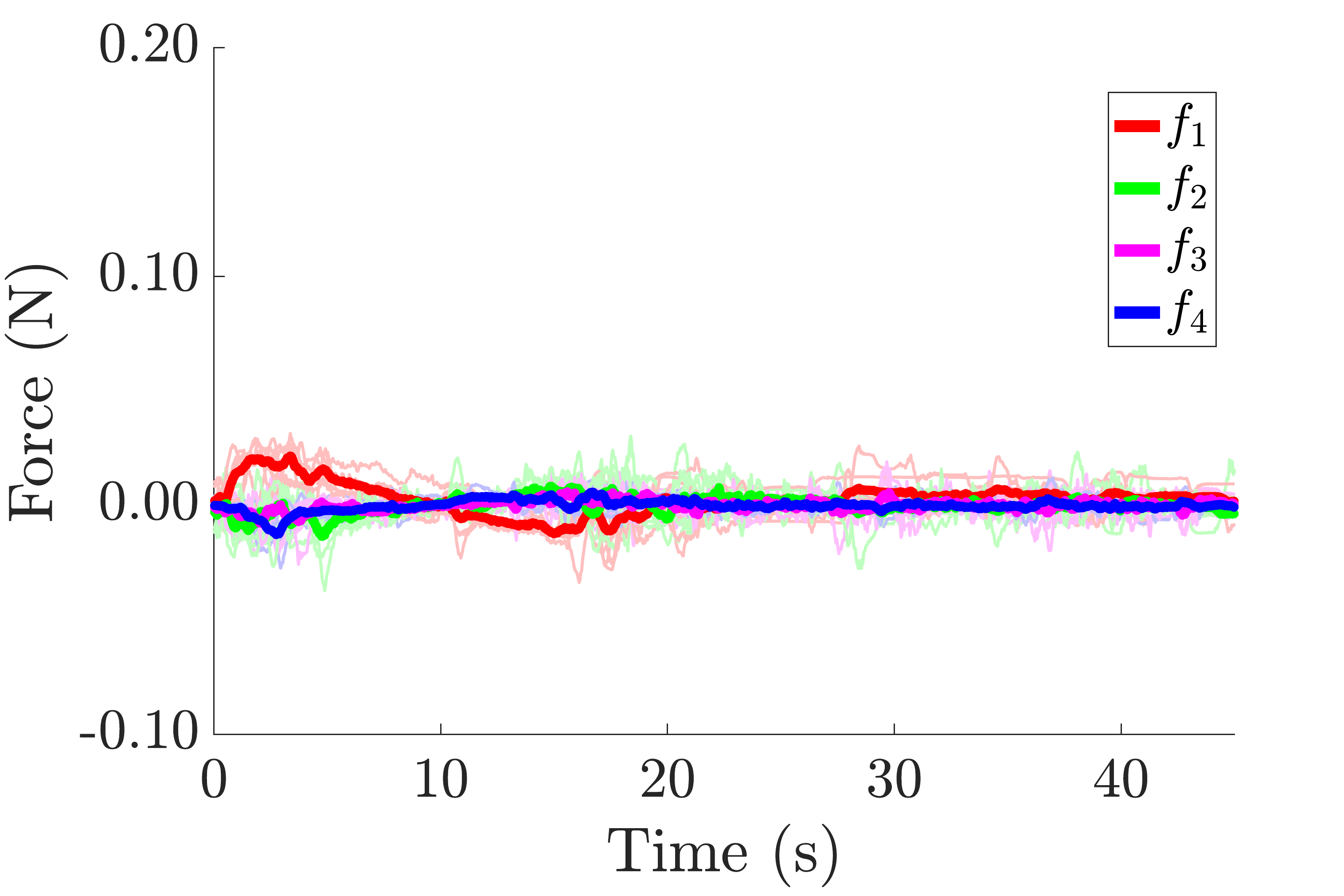}
  }
\subfloat[Deformations\label{fig_Deform_Plot_exp}]
  {
    \includegraphics[width=0.65\columnwidth]{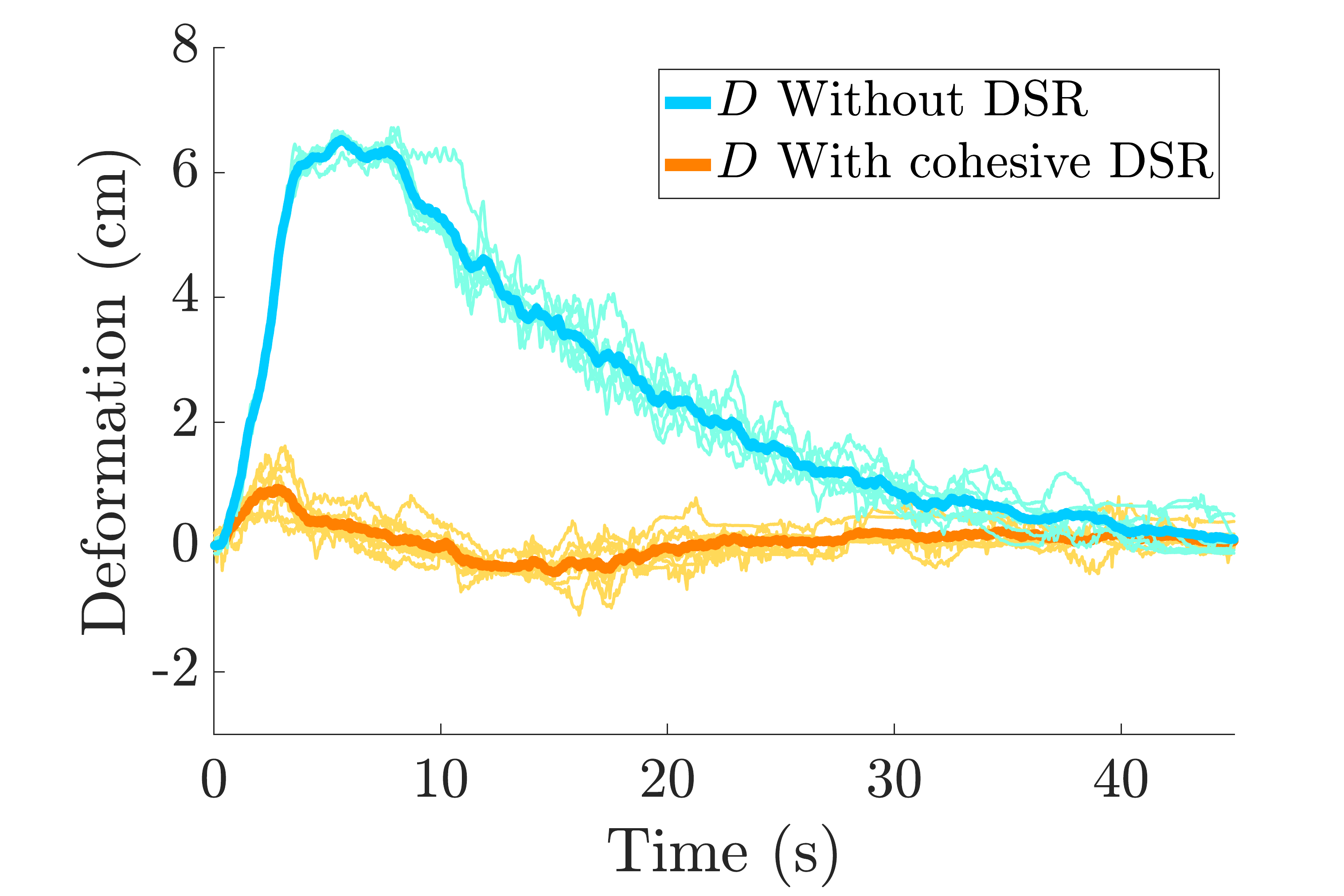}
  }
    \caption{Comparative evaluation of force $f_k$ as in Eq.~\eqref{Eq_force_single} and deformation  $D$ as in Eq.~\eqref{Eq_max_deformation} with and without cohesive DSR, and similarity of simulations (top row) and experimental results (bottom row).  
    Experiment results are shown for 7 trials (shown in thin lines), and the means are shown in thick lines.
    }
  \label{fig_result}
%  \vspace{-0.1in}
\end{figure*}

\section{Results}
\subsection{Selection of the desired transport  trajectory}
\begin{figure}[!ht]
  \centering
  \subfloat[ ]
  {
    \includegraphics[width=0.47\columnwidth]{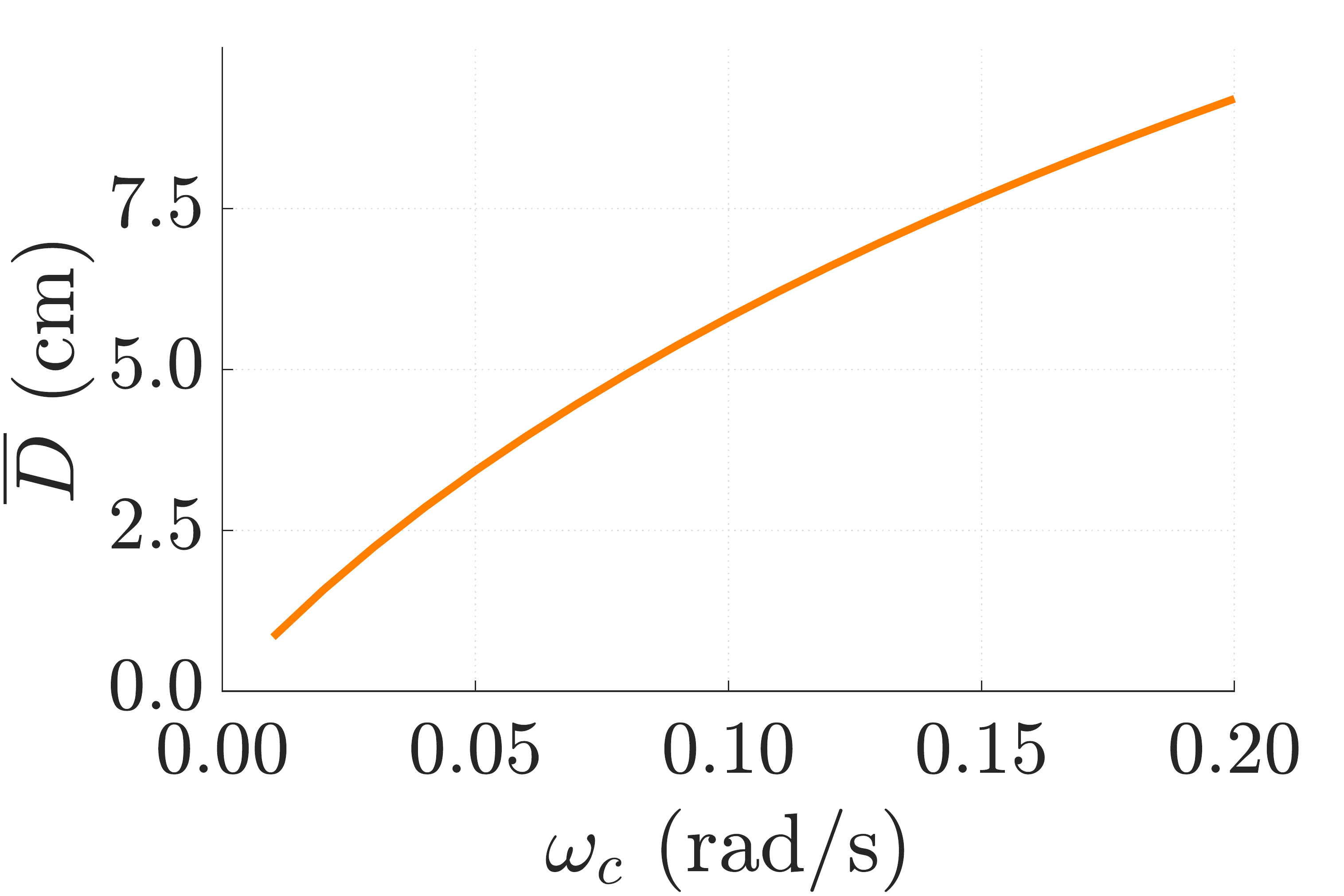}
    \label{wc_vs_Dbar}
  }
  \subfloat[ ]
  {
    \includegraphics[width=0.47\columnwidth]{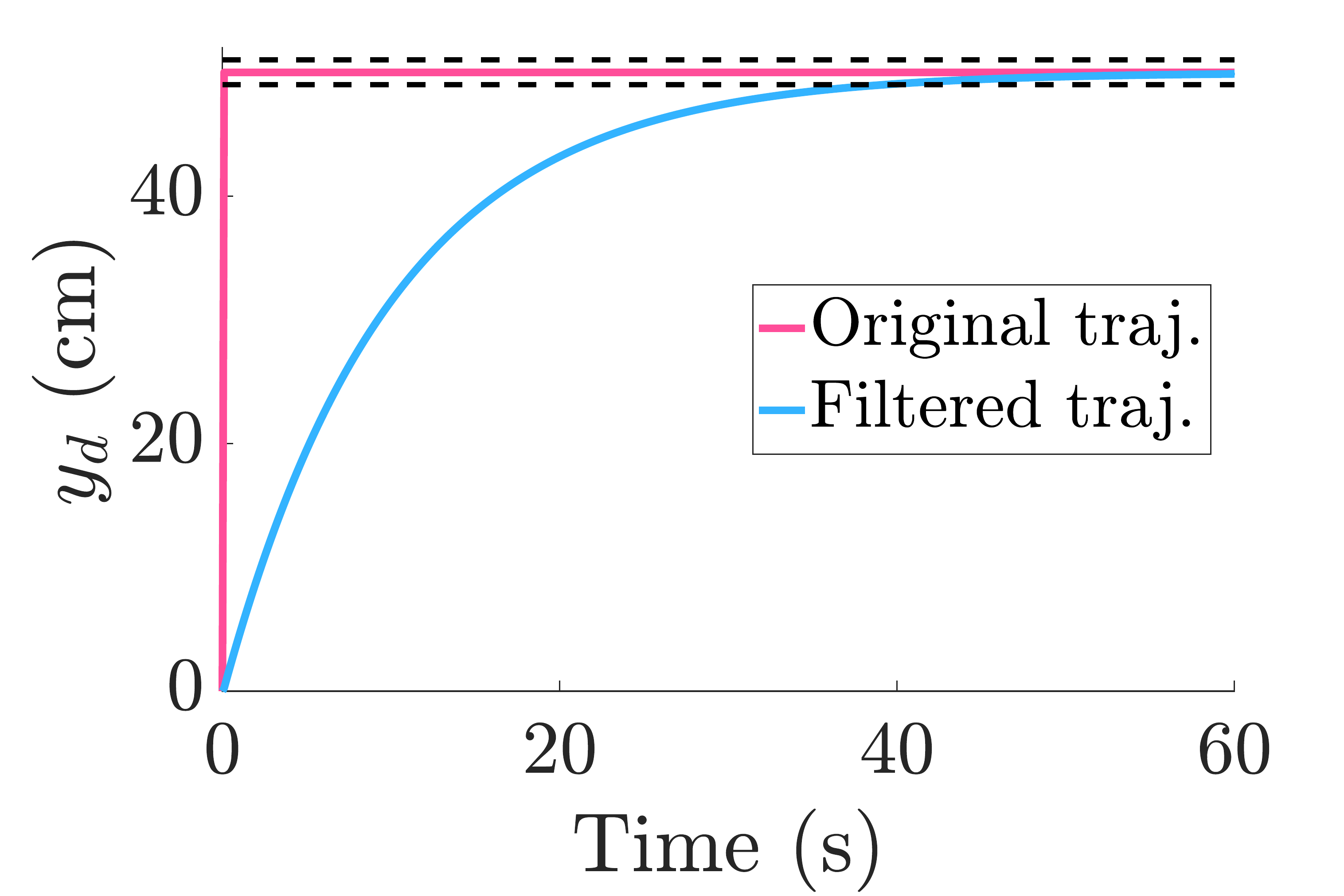}
    \label{fig_time_trajectory}
  }
  \caption{(a) The effect of cutoff frequency $\omega_c$ on maximum deformation $\overline{D}$. (b) The desired trajectory $y_d$ obtained by passing a step trajectory $y_{ds}$ through a  first-order, low-pass filter with cutoff frequency $\omega_c = 0.1 \; rad/s$ as in Eq.~\eqref{filter_traj_eq}. 
  }
  \label{freq_vs_d}
\end{figure}

A large change in position  $y_d$ from $0\;cm$ to $50\;cm$ was chosen to help visualize the transport of the flexible body. To ensure that the deformations are not too large (i.e., to avoid other robots dragging each other), the desired transport trajectory $y_d$ was chosen as a step that is filtered using a first-order, low-pass filter with cutoff frequency $\omega_c$ and implemented 
using Tustin's approximation, as
\begin{equation}
    y_{d}[m] = \frac{2 - \omega_c \delta_t}{2+ \omega_c \delta_t} y_{d}[m-1] + \frac{\omega_c \delta_t}{2+\omega_c \delta_t} \left( y_{ds}[m] + y_{ds}[m-1] \right),  
    \label{filter_traj_eq}
\end{equation}
where  $y_{ds}[m]=50$ if $m>0$ and zero otherwise.
The effect of the cutoff frequency $\omega_c$ on maximum deformation 
$\overline{D}$ in Eq.~\eqref{Eq_max_deformation} is shown in Fig.~\ref{wc_vs_Dbar}.
The cutoff frequency $\omega_c$ was selected as $\omega_c = 0.1 \; rad/s$ so that the maximum deformation $\overline{D}$ is below $7 \; cm$ and the maximum speed input to the robot $v_{d,k} \leq (v_{max} = 5\;cm/s)$ to avoid dragging of the robots by each other for the case without DSR. Note that the desired trajectory $y_d$ reaches the final value of $50 \; cm$ in about  $T_{sf} = 4/\omega_c = 40 \; s$ as seen in Fig.~\ref{fig_time_trajectory}.
\vspace{0.1in}

\subsection{Results and discussion}
\begin{figure}[ht!]
  \centering
  \subfloat[\label{fig_Force_Plot_without_DSR}]
  {
    \includegraphics[width=0.8\columnwidth]{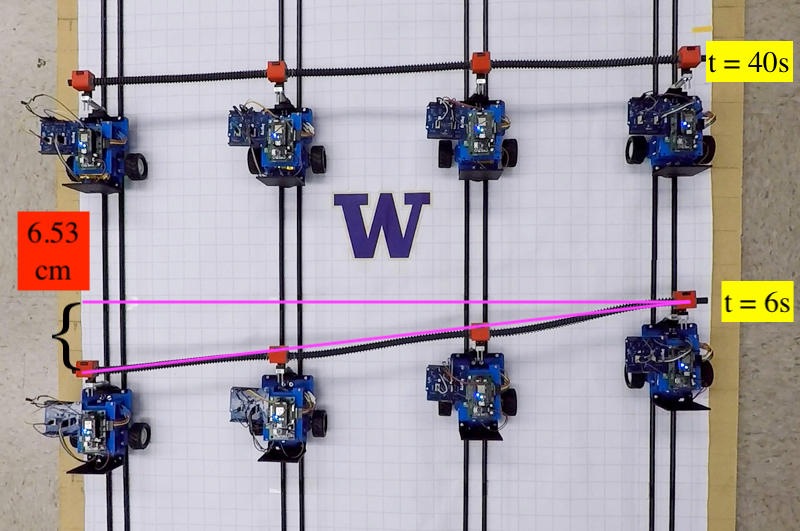}
  }\\
  \subfloat[\label{fig_Force_Plot_with_DSR}]
  {
    \includegraphics[width=0.8\columnwidth]{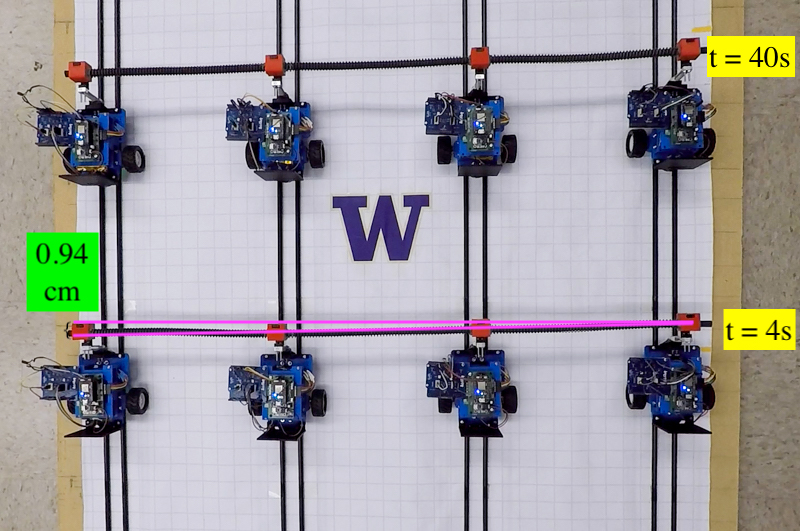}
  }
    \caption{Reduction in maximum deformation $\overline{D}$ with cohesive DSR approach (at time $t=4\;s$) compared to the case without DSR  (at time $t=6\;s$) as seen in video snapshots  of the experiment overlaid with the positions at time $t=40\;s$: (top) without DSR, and (bottom) with cohesive DSR. The deformations over time are shown in Fig.~\ref{fig_Deform_Plot_exp}. Video of the experiment can be seen here: \url{https://youtu.be/tzDfnMbgIgA}. 
    }
  \label{fig_snapshots}
\end{figure}

Comparative evaluations, with and without the DSR approach, are presented below. 
The results are evaluated based on the maximum deformation $\overline{D}$ in Eq.~\eqref{Eq_max_deformation} and also the  maximum force $\overline{f}$ defined as
    \begin{equation}
        \overline{f} = \max_{k=\{1,2,3,4\}} \left( \max_m |f_k[m]| \right).
        \label{Eq_max_force}
    \end{equation}
The simulation and experimental results are shown in Fig.~\ref{fig_result}, and quantified in  Table ~\ref{result_summary}.
The responses from the experiments and simulations are similar to each other in Fig.~\ref{fig_result}, which indicates that the models are close to the experimental system.  
The  cohesive DSR approach
reduces the maximum deformation  $\overline{D}$ substantially, by $90\%$ in simulation and $85\pm0.05\%$ in experiment. Similarly, the corresponding maximum forces $\overline{f}$ are also reduced  significantly, by $90\%$ in simulation and $87\pm0.50\%$ in experiment. The reduction in deformation indicates that the robot network responses are more cohesive during transport with the cohesive DSR approach. This increased cohesion can also be observed from  snapshots of experiment in Fig.~\ref{fig_snapshots}.
% {\color{blue} Moreover, the cohesive DSR reduces the forces close to the noise level of the sensor as shown in Fig.~\ref{noise_level}.} 
%
\begin{table}[!ht]
\begin{center}
    \begin{tabular}{ c | c  c  c   } 
    \hline
    Label & Without DSR & Cohesive DSR & Improvement \\
    \hline\hline
    \multicolumn{4}{c}{\cellcolor{cyan!25} \textbf{Simulation}}\\
    \hline
    $\overline{f}$ (N) & 0.146 & 0.014 & 90\% \\

    \hline
    $\overline{D}$ (cm) & 5.824 & 0.563 & 90\% \\

    \hline
    
    \hline\hline
    \multicolumn{4}{c}{\cellcolor{cyan!25}\textbf{Experiment}}\\
    \hline
    $\overline{f}$ ($\mu \pm \sigma$) (N) & 0.165 $\pm$0.008  & 0.021 $\pm$0.004  & 87 $\pm$0.50\%  \\
    \hline
    $\overline{D}$ ($\mu \pm \sigma$) (cm) & 6.530 $\pm$0.010 & 0.940 $\pm$0.010  & 85 $\pm$0.05\% \\
    \hline
    \end{tabular}
 \caption{Improvement (reduction) in maximum force ( $\overline{f}$) and maximum deformation ($\overline{D}$) with cohesive DSR when compared to case without DSR. 
    Top: Simulation results. 
    Bottom: Experimental results with mean $\mu$ and standard deviation $\sigma$ over 7 trials.
    }
    \label{result_summary}
\end{center} 
\end{table}
\vspace{0.5in}
\section{Conclusion}
An approach was presented to reduce deformation of objects during transport with decentralized robot networks. The approach used only local force measurements without additional communication, and conditions for stability were established.  The proposed cohesive DSR approach was evaluated using simulation and the results closely matched the experimental results. 
 Overall, the proposed approach led to $85\%$ reduction in the deformation of the experimental system without increasing the time to transport the object to a new position. Ongoing efforts are focused on extending the approach to systems with high-order dynamics. 
%
%
%

%%%%%%%%%%%%%%%%%%%%%%%%%%%%%%%%%%%%%%%%%%%%%%%%%%%%%%%%%%%%%%%%%%%%%%%%%%%%%%%%

% \addtolength{\textheight}{-12cm}   % This command serves to balance the column lengths
                                  % on the last page of the document manually. It shortens
                                  % the textheight of the last page by a suitable amount.
                                  % This command does not take effect until the next page
                                  % so it should come on the page before the last. Make
                                  % sure that you do not shorten the textheight too much.

%%%%%%%%%%%%%%%%%%%%%%%%%%%%%%%%%%%%%%%%%%%%%%%%%%%%%%%%%%%%%%%%%%%%%%%%%%%%%%%%

%%%%%%%%%%%%%%%%%%%%%%%%%%%%%%%%%%%%%%%%%%%%%%%%%%%%%%%%%%%%%%%%%%%%%%%%%%%%%%%%

%%%%%%%%%%%%%%%%%%%%%%%%%%%%%%%%%%%%%%%%%%%%%%%%%%%%%%%%%%%%%%%%%%%%%%%%%%%%%%%%
% \section*{APPENDIX}

% Appendixes should appear before the acknowledgment.

% \section*{ACKNOWLEDGMENT}

%%%%%%%%%%%%%%%%%%%%%%%%%%%%%%%%%%%%%%%%%%%%%%%%%%%%%%%%%%%%%%%%%%%%%%%%%%%%%%%%
% \newpage
\bibliographystyle{IEEEtran}
\bibliography{mybib}

\end{document}